\documentclass[conference]{IEEEtran}
\IEEEoverridecommandlockouts
\usepackage{cite}
\usepackage{amsmath,amssymb,amsfonts}
\usepackage{algorithmic}
\usepackage{graphicx}
\usepackage{textcomp}
\usepackage{xcolor}
\usepackage{hyperref}

\usepackage{subcaption}
\usepackage{booktabs}
\usepackage[ruled,vlined]{algorithm2e}

\def\BibTeX{{\rm B\kern-.05em{\sc i\kern-.025em b}\kern-.08em
    T\kern-.1667em\lower.7ex\hbox{E}\kern-.125emX}}
\begin{document}

\newcommand{\concat}{%
  \mathbin{{+}\mspace{-8mu}{+}}%
}

\title{A Multimodal Learning-based Approach for Autonomous Landing of UAV
}

\author{\IEEEauthorblockN{Francisco S. Neves, Luís M. Branco, Maria I. Pereira, Rafael M. Claro, Andry M. Pinto}
\IEEEauthorblockA{\textit{Department of Electrical and Computer Engineering (DEEC), Faculty of Engineering of University of Porto (FEUP)} \\
\textit{Institute for Systems and Computer Engineering - Technology and Science (INESC TEC), Porto, Portugal}\\
\{francisco.s.pinto, maria.i.pereira, rafael.marques\}@inesctec.pt, \{up201906466, amgp\}@fe.up.pt}
}

\maketitle

\begin{abstract}
In the field of autonomous Unmanned Aerial Vehicles (UAVs) landing, conventional approaches fall short in delivering not only the required precision but also the resilience against environmental disturbances. Yet, learning-based algorithms can offer promising solutions by leveraging their ability to learn the intelligent behaviour from data. On one hand, this paper introduces a novel multimodal transformer-based Deep Learning detector, that can provide reliable positioning for precise autonomous landing. It surpasses standard approaches by addressing individual sensor limitations, achieving high reliability even in diverse weather and sensor failure conditions. It was rigorously validated across varying environments, achieving optimal true positive rates and average precisions of up to 90\%. On the other hand, it is proposed a Reinforcement Learning (RL) decision-making model, based on a Deep Q-Network (DQN) rationale. Initially trained in sumlation, its adaptive behaviour is successfully transferred and validated in a real outdoor scenario. Furthermore, this approach demonstrates rapid inference times of approximately 5ms, validating its applicability on edge devices.

\end{abstract}


\section{Introduction}
UAV technology advancements have allowed for greater incorporation of these vehicles into the real-world in a broad set of applications such as: including civil transportation \cite{moshref2021applications,biccici2021approach}, monitoring and surveying \cite{zhao2021structural}, search and rescue \cite{martinez2021search}, and specifically infrastructure inspection \cite{falorca2021new}, due to their ability to explore the unstructured environment and employ versatile movements. Moreover, there is a growing demand for enhanced autonomy, particularly as the complexity of the application increases and necessitates UAVs to operate Beyond Visual Line of Sight (BVLOS) locations. For example, in situations like offshore missions, it is crucial to implement fully autonomous UAV policies, thus eliminating the need for human intervention, reducing potential risks \cite{pinto2021atlantis}.

The effectiveness of an autonomous UAV landing procedure is reliant, on one hand, on the precision of the landing target detection, and on the other hand on the ability to produce the right succession of maneuvers to guide and land UAV on the platform. Furthermore, because these operations must be conducted under challenging environmental disturbances such as varying daytime conditions, rain, fog, and uncertain lighting, UAVs require multiple sensors that complement each other's limitations, and decision-making policies that are able to adapt to such hectic conditions.

A precise landing with centimeter-level precision can not be provided by GPS sensors \cite{b2}. An alternative involves endowing landing platforms with artificial markers detectable by visual \cite{garrido2014automatic} or thermographic \cite{khattak2018marker} cameras using computer-vision methods ~\cite{rabah2018autonomous,safadinho2020uav}. Yet, these solutions become unfeasible in adverse lighting conditions \cite{davis2019reflective}, which particularly affect these sensors. Also, because these markers lack depth information, it becomes necessary to rely on 3D range sensors such as LiDAR (Light Detection And Ranging), capable of recognizing artificial markers with retro-reflective properties \cite{davis2019reflective}. Therefore, it is clear that depending solely on individual sensors is susceptible to be compromised under specific weather conditions. And thus, because unimodal approaches are subjected to particular disadvantages, it is crucial to resort to multimodal solutions. Recently, Claro et al., proposed ArTuga ~\cite{claro2023artuga}, a fiducial multimodal-based marker detectable by visual, thermographic, and LiDAR sensors. Such a multimodal marker enables multimodal detection for an UAV equipped with an heterogeneous perception system containing these sensors. And so, with such a perception system that uses several complementary modalities with centimeter-level accuracy, a decision-making algorithm has superior conditions to generate landing maneuvers that are more reliable and precise.

Adverse weather conditions have the potential to introduce artifacts, occlusions, and noise, thereby compromising the integrity of acquired sensory information. And under such circumstances, conventional geometric computer vision methodologies that rely on this information encounter limitations. Existent solutions resort to Deep Learning, because they can adapt to inherently incomplete information. This is achieved by learning to generalize to diverse input conditions~\cite{lecun2015deep}. The progress made in this field over the past ten years has been specially fueled by the CNNs' ongoing success in a variety of perception-related tasks~\cite{lecun2010convolutional}. Nonetheless, more recently, there has been an increased focus on Transformer-based neural network architectures ~\cite{vaswani2017attention}, initially recognized for their outstanding performance in natural language processing tasks and more recently extended to visual perception~\cite{dosovitskiy2020image}. Moreover, owing to their inherent attention mechanisms, transformers exhibit a natural capability for cross-modality processing across multiple signals~\cite{nagrani2021attention}. Hence, visual transformers can also address multimodal objectives.

One common approach to performing navigation and landing involves complex mathematical methods that may prove too intricate and limiting in unpredictable environments. Traditional approaches ~\cite{UAV_complexmath} rely heavily on PID controllers, Kalman Filters, and pose estimation using inertial sensors. However, utilizing such complex techniques could pose challenges in effectively handling unexpected or dynamic situations, potentially compromising the overall effectiveness and safety of the navigation system.

In the pursuit of achieving full autonomy in navigation, researchers have turned their attention to integrating RL as a promising approach for improvement~\cite{UAV_highmodel}. 
By not requiring a model and due to its straightforwad formulation, one of the most promising RL algorithms is Q-Learning \cite{q_learning}. It is a method in Reinforcement Learning where an agent learns to make optimal decisions over time. It iteratively updates its understanding of the best actions to take in different situations, represented by the action-value function $Q$. For each state $s$ selects the action $a$ that maximizes the expected return. At time $t$, the agent will be given a reward $r$ after taking an action $a$ in state $s$.
Q-values are updated iteratively using the following formula until convergence or termination:
    \begin{equation}
    Q(s_t, a_t) \leftarrow (1-\alpha)Q(s_t, a_t) + \alpha (r_{t+1} + \gamma \max_{a_{t+1}} Q(s_{t+1}, a_{t+1}))
    \end{equation}
where \(\alpha\) and \(\gamma\) are the learning rate and discount rate, respectively. Moreover, Deep Q-Network (DQN), first proposed by ~\cite{Dqn_atari} is an algorithm in the field of RL that combines the principles of deep neural networks with Q-learning, enabling agents to learn optimal policies in complex environments. DQN uses a deep neural network to estimate the Q-values for each (state, action) pair. The neural network takes the state as input and outputs the Q-value for each action. These values represent the expected cumulative reward of executing the action and following the optimal policy thereafter.

Given the real-time nature of this application, it is crucial to select a decision-making model that can promptly process data. Also, it should have minimal complexity to be easily accomodated by the limited embedded resources. As such, there is a balance between simplicity and the preservation of the decision-making capacity. And since the model will operate autonomously onboard the UAV, it is essential for it to be easily interpretable for effective monitoring by a human supervisor since legal constraints do not allow BVLOS operations. As a result, this design due to this limited number of states and actions ensures a faster convergence which in turn leads to a more stable behaviour. To address these requirements, it is proposed a discretized version of a DQN model. This adaptation of DQN provides the interpretability of a discrete RL model, like Q-learning, while leveraging its neural network nature to generalize and reason effectively in response to hazard environmental scenarios. Therefore, this study introduces an innovative precise landing system that integrates both a multimodal transformer-based perception system with a deep Q-learning reinforcement-learning decision-making system. This system demonstrates reliability across distinct lighting, smoke and fog conditions to produce the right set of autonomous maneuvers to precisely land. The contributions of this research are the following:

\begin{enumerate}
    \item A novel multimodal transformer-based Deep Learning neural network, specifically designed to enhance the reliability of the landing target detection capabilities of UAVs, under challenging environmental conditions.

    \item An autonomous decision-making deep reinforcement-learning algorithm that learns to execute the right set of actions for precise landing of an UAV;

    \item An embedded deployable and integrated autonomous perception and decision-making framework validated on modern edge devices;
    
    
\end{enumerate}

This paper is structured as follows: Section \ref{sec:method} presents the proposed the precise landing system by describing each of its components such as the landing target detector, and the reinforcement learning lander. Section \ref{sec:results} demonstrates the conducted training and real-world validation experiments. Section \ref{sec:conclusions} summarizes the most important conclusions.

\section{Learning-based Autonomous Lander}\label{sec:method}
This section introduces the two modules of the learning-based autonomous lander. Firstly, in section \ref{subsec:detector}, it is presented the landing target detector. This module is a perception system that learns to process multimodal information such as visual, thermal, and LiDAR data to determine the position of the landing target within the visual camera frame. Secondly, in section \ref{subsec:rl_lander}, we discuss the reinforcement-learning lander. This module reads the landing target position provided by the perception module and outputs the next waypoint position for the UAV to navigate towards.

\subsection{Multimodal Landing Target Detector}\label{subsec:detector}
\begin{figure}[htbp]
\centering
\includegraphics[width=\linewidth]{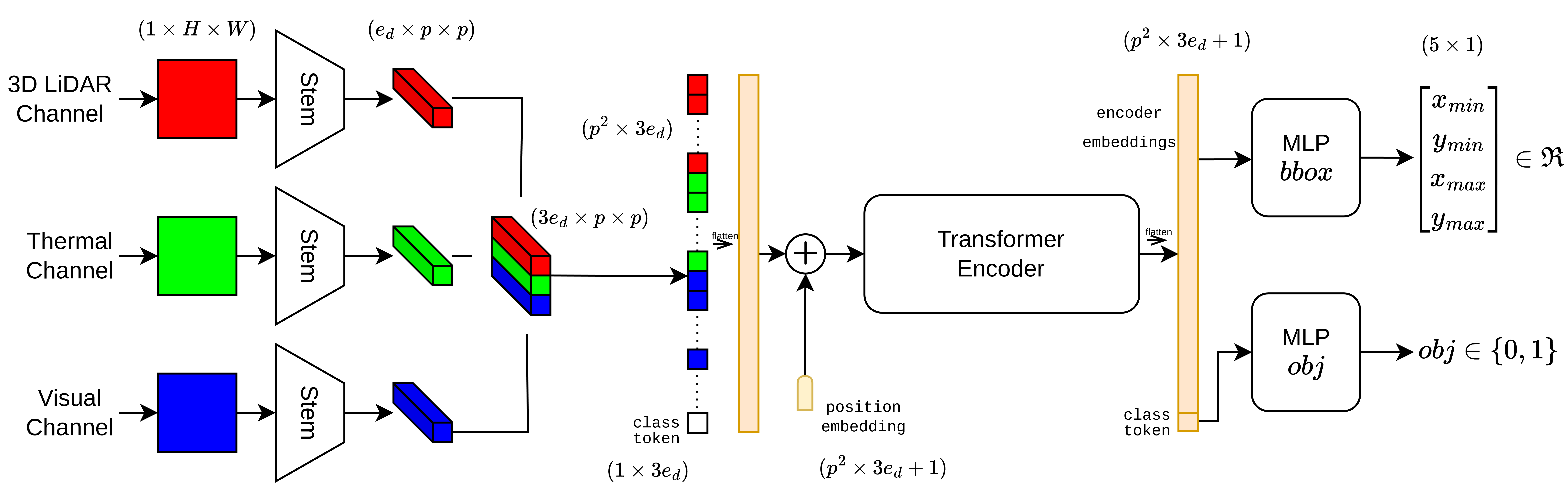}
\caption{The multimodal landing target detector architecture.}
\label{fig:vital_architecture}
\end{figure}

ViTAL (Vision Transformer for Artuga-based Landing) is a multimodal fusion detection model. ViTAL improves upon the traditional Vision Transformer (ViT)~\cite{b2} by integrating multimodal fusion, and adding image localization through bounding box detection. The overall ViTAL architecture contains three main parts: as follows: a backbone containing a block of CNNs for feature extraction, a transformer encoder to produce the embeddings, and a prediction block composed of two multi-layer perceptrons (MLP) heads to generate the final prediction, as depicted in the Figure \ref{fig:vital_architecture}. The prediction is comprised by an $objectness \in [0,1]$ score, representing the probability the marker object is present in the input image, and a bounding box ($bbox=\begin{bmatrix}x_{min} & y_{min} & x_{max} & y_{max}\end{bmatrix}$) vector, where $(x_{min}, y_{min})$ and $(x_{max}, y_{max})$ are the coordinates of the top-left and the bottom-right corners of the rectangular 2D box enclosing the detected object, respectively.

The dataset is designed according to the data preparation methodology employed in previous work~\cite{neves2023end}, using distinct sensory data from visual, thermal, and LiDAR sensors. Several pre-processing and calibration techniques are applied to the raw data to generate a concatenated RGB representation. In this representation, each red, green, and blue channel corresponds to an individual grayscale image containing visual, thermal, and LiDAR information, respectively. This concatenated image is subsequently split into separate channels to be utilized as input for ViTAL. Additionally, each image has a corresponding label containing a manually annotated bounding box identifying the ArTuga, along with a binary classification token indicating the presence of the marker in the image.


\subsubsection{Architecture}
\paragraph{Backbone} The backbone of ViTAL consists of three convolutional stems and additional geometric operations essential for preprocessing the input for the transformer encoder. These stems learn to extract relevant features from the input image, generating latent fixed-size patches that are subsequently fed into the transformer encoder. A grayscale image $x_{img} \in \Re^{1\times H \times W}$, representative of each modality, feeds each stem, where $H$ and $W$ represent the height and width, respectively. According to \cite{xiao2021early}, applying early convolutional stems, rather than conventional patchifying stems as in the original ViT framework, enhances the transformer's understanding of images, thus improving training convergence and stability. Each stem consists of three consecutive blocks, with each block sequentially processing its input through two 3x3 convolutions with a stride of 1 and padding of 1. Following each convolutional operation, a batch normalization layer is applied, and a ReLU activation function is inserted between the convolutions. Surrounding each block, there is a residual connection where the input goes through a 3x3 convolutional operation with a stride of 1 and padding of 0. At the end of each block, the output is added to the residual, followed by a ReLU activation and a max pooling operation, ultimately reducing the input size by half. Finally, the output from the last stem block undergoes a final 3x3 convolutional operation with a stride of 1 and padding of 1. Each stem produces an amount of $e_d=128$\footnote{$e_d$ stands for the dimension of the embeddings.} patches $x_p \in e_d \times p \times p$, where $p=20$ is the side length of each patch. Next, the outputs from the three stems are concatenated along the $e_d$ dimension. The resulting concatenated output is then flattened to attain a final size of $x_{flat} \in \Re^{p^2 \times 3e_d}$. Subsequently, the flattened output is concatenated with a trainable embedding class token $x_{cls} \in \Re^{1 \times 3e_d}$ along the $p^2$ dimension. Finally, to integrate positional information, the concatenated output ($x_{flat} \concat x_{cls}$) is added to a trainable positional embedding $x_{pos} \in \Re^{(p^2 + 1) \times e_d}$, ultimately serving as the input embedding to the transformer encoder.

\paragraph{Transformer Encoder} This module involves understanding and capturing complex relationships from the input data. It follows the structure proposed by \cite{vaswani2017attention}. It stacks {$N_{enc}$} consecutive layers comprised by multi-head attention and feed-forward blocks. ViTAL has $N_{enc} = 6$, and it has 512 hidden dimensions produced by its encoder layers.

\paragraph{MLP heads} ViTAL employs two Multi-Layer Perceptron (MLP) heads, referred to as $MLP_{obj}$ and $MLP_{box}$. Both receive, as input, the encoded information provided by the transformer to generate the following outputs: the likelihood of object presence and determining; and the bounding box encapsulating the identified marker, respectively. Each head consists of a 2-layer architecture: a hidden layer with $e_d$ neurons incorporating layer normalization and a GELU activation function; and a linear layer with layer normalization, comprising $4$ neurons for the $MLP_{box}$ and $1$ neuron for the $MLP_{obj}$. A dropout regularization with a probability of $0.2$ is applied between the layers.

\subsubsection{Loss function} 
The loss function of this model consists of two components designed to tackle two distinct challenges: localization and classification. The bounding box head, which is responsible for identifying the bounding box containing the landing marker, employs the Complete IoU (CIoU) loss function~\cite{zheng2021enhancing}. CIoU represents a state-of-the-art approach to bounding box regression, taking into consideration factors such as aspect ratio, area overlap, and the distance between predicted and ground-truth bounding boxes during the learning process. CIoU is chosen for its slightly superior performance against similar purpose counterparts like DIoU~\cite{zheng2020distance} or GIoU~\cite{Rezatofighi_2018_CVPR}, demonstrating the best results in empirical evaluations as exposed in the Table \ref{tab:losses}. On the other hand, the objectness head, tasked with determining whether each image contains the object of interest, utilizes the Focal loss function~\cite{lin2017focal}. The Focal Loss function addresses class imbalance by incorporating an additional modulating term into the standard cross-entropy criteria.  Given the substantial class imbalance between positive and negative samples in the dataset ($85\%$ to $15\%$, respectively), Focal Loss was selected over standard Binary Cross Entropy loss, yielding also enhanced outcomes due to the inherent ability of handling class imbalance.

\begin{table}[htbp]
    \caption{Effect of the different bounding box losses on training performance. $AP_{50}$ and $AP_{50:95}$ are the average precisions for IoU thresholds of $50\%$ and from $50\%$ to $95\%$, respectively.}
    \centering
    \small
    \begin{tabular}{rcc}
         \textbf{Loss Function} & $AP_{50}$ & $AP_{50:95}$ \\
         \hline
         GIoU & $0.852$ & $0.389$ \\
         DIoU & $0.907$ & $0.441$ \\
         CIoU & $0.896$ & $\textbf{0.453}$ \\
    \end{tabular}
    \label{tab:losses}
\end{table}

\subsection{Reinforcement Learning Decision-Making Lander}\label{subsec:rl_lander}

The DQN architecture proposed by~\cite{Dqn_atari} comprises a sequence of layers designed to process state information and approximate action values efficiently. It takes raw pixel data from the game screen as input, processes it through convolutional layers to extract features, and then passes it through fully connected layers to estimate Q-values for each possible action.
In this particular work, the architecture does not include convolutional layers as images are not being used as input, but rather an unidimensional vector state. Hence, the model only includes fully connected layers. More precisely, it starts with an input layer comprising three neurons, each representing a component of the state representation. It is followed by three intermediate layers, each containing 256, 256, and 128 neurons, respectively. Finally, the network concludes with an output layer of five neurons, representing the available actions that the agent can take in the environment.

\begin{figure}[htbp]
\centering
\includegraphics[scale=0.35]{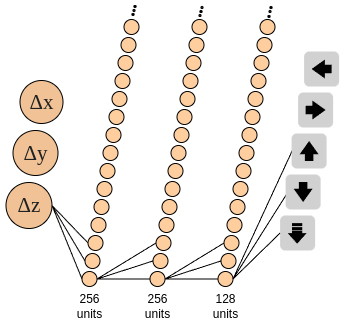}
\caption{The RL decision-making lander architecture.}
\label{fig:dqn_architecture}
\end{figure}

\subsubsection{Problem Formulation}
The objective is to autonomously achieve a precise landing of a UAV onto a designated landing pad. By analyzing the disparity between the landing pad position and UAV relative position, the UAV can execute actions to minimize this difference, thereby refining its trajectory toward the landing pad. The available actions for the UAV are classified as moving forward, backward, left, right, or descending vertically. Crucially, each action preserves the UAV's orientation throughout the trajectory.

In our pursuit of achieving autonomous precision landing for UAV, we employ a reinforcement learning framework, encompassing the environment, state space, action space, and reward function.
The environment constitutes the spatial domain surrounding the designated landing pad. It encapsulates the physical space through which the UAV navigates during the landing task.
The environment was discretized into a 3D grid with a resolution of 1 meter. The spatial extents of the environment span from -6 to 6 meters along the x and y axes, and 8 meters along the z axis. Discretizing the space allows for enhanced interpretability of the data and model outputs. By dividing the continuous space, it becomes easier to understand and analyze the relationships within the data. Additionally, discretization helps ensure convergence in computational processes.
The state $s$ of the environment is represented by a three-dimensional vector $(\Delta x, \Delta y, \Delta z)$, where $\Delta x$ and $\Delta y$ denote the horizontal relative position coordinates of the UAV relative to the landing pad, and $\Delta z$ represents the height of the UAV above the ground level measured by the UAV's onboard altimeter. Given the coordinates of the bounding box $(x_{min}, y_{min}, x_{max}, y_{max}$), provided by the detector, the center point of the image $(c_x, c_y)$, $\Delta x$ and $\Delta y$ are obtained as follows:
\begin{align}
    \Delta x = \frac{x_{max}-x_{min}}{2} - c_x, \quad
    \Delta y = \frac{y_{max}-y_{min}}{2} - c_y.
\end{align}

The action space comprises a discrete set of actions available to the UAV, including moving forward, backward, left, right, and descending vertically. Each action enables the UAV to dynamically adjust its position and altitude relative to the landing pad, facilitating trajectory optimization.

The reward function serves as a critical component of the RL framework, providing feedback to the UAV based on its actions and current state. It incorporates a shaping mechanism that aims to encourage the UAV to approach the landing pad. 

\begin{equation}
shapping_t = -100 \times \sqrt{K_x \cdot (\Delta_x)^2 + K_y \cdot (\Delta_y)^2 + K_z \cdot (\Delta_z)^2}
\end{equation}

The shaping calculation has a constant multiplied by the relative position to the landing pad position. The shaping of the reward varies depending on whether the UAV is inside a specified region around the landing pad. If the UAV is already inside this region, the reward shaping primarily focuses on the altitude component. However, if the UAV is outside this region, the reward shaping incorporates both the horizontal position and altitude components. 
The weights K allow for a more controlled adjustment of the shaping, ensuring that the behavior is influenced appropriately based on its position relative to the landing pad.

The reward is adjusted based on the difference between the current shaping and the previous shaping:
\begin{equation}
reward_t = shaping_t - shaping_{t-1}
\end{equation}

If the UAV has successfully landed, it receives a reward of 400, indicating a successful completion of the landing task. However, if the UAV has crashed, the reward is calculated as -200 multiplied by the distance to the landing pad. This penalty encourages the UAV to avoid crashing. 



\begin{algorithm}
\SetAlgoLined
\eIf{UAV is inside Landing Zone}
{
    \eIf{UAV was inside at previous state}
    {
        {$shaping \gets r_{land}$}\;
        {$next\_shaping \gets shaping$}\;
    }
    {
        {$shaping \gets r_{appr}$}\;
        {$next\_shaping \gets r_{land}$}\;
    }
}
{
    \If{UAV left Landing Zone}
    {
        {$prev\_shaping \gets r_{appr}$}\;
    }
    {$shaping \gets r_{appr}$}\;
    {$next\_shaping \gets shaping$}\;
}

{$reward \gets shaping - prev\_shaping$}\;
{$prev\_shaping \gets next\_shaping$}\;

\If{UAV landed successfully}
    {
        {$reward \gets 400$}\;
    }
\If{UAV landed outside Landing Zone}
    {
        {$reward \gets -200 \times distance\ to\ pad$}\;
    }

 \caption{Reward Function}
 
%
%

\end{algorithm}


The training process for the UAV landing model revolves around the Deep Q-Network algorithm. This iterative process entails engaging with the environment, accumulating experiences, and adjusting neural network parameters to refine the policy. During training, the agent employs an epsilon-greedy algorithm to choose actions based on its current state, ensuring a balance between exploration and exploitation.  The incorporation of periodic soft updates to the target network, as proposed by ~\cite{soft_DDPG}, significantly stabilizes the learning dynamics, enhancing convergence and overall training efficiency.
A stopping criterion is evaluated based on the minimal and mean returns over a certain number of episodes. Training is considered complete if the stopping criterion is met or if the maximum number of episodes is reached.

\begin{table}[htbp]
    \caption{Hyperparameters for training.}
    \centering
    \small
    \begin{tabular}{c|c}
        \textbf{Hyperparameter} & \textbf{Value} \\
        \hline
        Episodes (E) & 5000 \\
        Initial Epsilon (\(\varepsilon_i\)) & 1.0 \\
        Final Epsilon (\(\varepsilon_f\)) & 0.1 \\
        Decrease of Epsilon (\(\Delta\varepsilon\)) & 0.005 \\
        Discount Factor (\(\gamma\)) & 0.99 \\
        Learning Rate (\(\alpha\)) & 0.001 \\
        Target Net Update (\(\tau\)) & 0.01 \\
        Batch size & 32 \\
    \end{tabular}
    \label{tab:RL_hyperparameters}
\end{table}

\section{Results}\label{sec:results}
This section presents the training results, and further real-world evaluations conducted in an outdoor setting to validate the performance of both the landing target detector in the section \ref{subsec:detector_results} and the precise lander modules \ref{subsec:rl_lander_results}. Each module is validated separately, which means the RL lander relies on RTK GPS to obtain its state. Furthermore, given the negligible inference time of the RL module, on the order of nanoseconds, as opposed to the detector which operates on the scale of milliseconds, only the inference time results for the detector are presented. The robotic platform used for the experiments is CROW, an UAV with a quadcopter frame developed by INESC TEC \cite{claro2023energy}. 
    
\subsection{Multimodal Landing Target Detector}\label{subsec:detector_results}
\subsubsection{Training}
\begin{figure}[htbp]
\centering
\includegraphics[width=\linewidth]{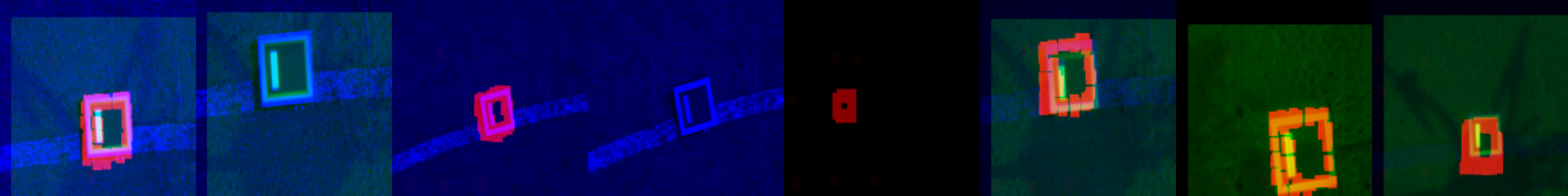}
\caption{Examples of dataset samples with various altitude, positioning and modality activation cases.}
\label{fig:dataset_samples}
\end{figure}

The detector trained on a dataset containing 914 3-channel images, which were divided into training, testing, and validation sets with proportions of $80\%$, $10\%$, and $10\%$, respectively. Each sample in the dataset is represented by a 5-length tuple that includes bounding box coordinates and a binary indicator representing the presence of a landing marker, termed $objectness$, structured as $(x_{min},y_{min},x_{max},y_{max},objectness)$. To enhance generalization and optimize training performance, augmentation techniques were applied, expanding the dataset to a final size of 6890 samples. These techniques included random horizontal and vertical flips with a probability of $50\%$, along with the introduction of salt and pepper noise with a probability of $0.2\%$. About $15\%$ of the dataset is made up of negative samples. Those samples are used to train the detector in situations where the marker is absent. Although an imbalanced ratio of negative to positive samples is expected to lead to a biased training, this impact is mitigated by the incorporation of the Focal Loss function throughout the training procedure.
Figure \ref{fig:dataset_samples} illustrates a few examples from the dataset.

\begin{table}[htbp]
    \centering
    \small
    \caption{ViTAL's training hyperparameters.}
    \label{tab:train_hyperparameters}
    \begin{tabular}{c|c}
         \textbf{Hyperparameter} & \textbf{Value} \\
         \hline
         Epochs & $200$ \\
         Learning rate & $3\times10^{-4}$ \\
         Batch size & $16$ \\
         Patience  & $30$ \\
         Weight decay & $0.01$ \\
    \end{tabular}
\end{table}
The training was conducted on an NVIDIA GeForce RTX 3070 GPU, an AMD Ryzen 7 5800H CPU x 16, 16GB of RAM, and a 64-bit Ubuntu 20.04 operating system. It were chosen a batch size of 16, and 200 epochs along with a corresponding Early Stopping strategy with a patience of 30 epochs. In accordance with the standard ViT setup, a low base learning rate of $3\times 10^{-4}$ was chosen. A reduce learning on plateau strategy was then implemented, monitoring the validation loss, with a factor of 0.1, a minimum learning rate of $10^{-6}$, and a threshold of 0.05. An AdamW optimizer with $\beta_1=0.9$, $\beta_2=0.999$, $\epsilon=10^{-8}$, and a weight decay of 0.01 was selected.
The selected hyperparameters are exposed in the Table \ref{tab:train_hyperparameters}.



For evaluating model performance, a weighted binary accuracy ($acc_{obj}$) was computed after each epoch to assess the objectness's performance, using a threshold of $0.5$ as follows:

\begin{equation}
acc_{obj} = \alpha\frac{TP}{TP+FN} + (1-\alpha)\frac{TN}{TN+FP},
\end{equation}
where $\alpha \in [0,1]$ denotes the balancing factor between positive and negative samples in the dataset. Conversely, the average IoU was employed to measure bounding box localization accuracy, defined as: 
\begin{equation}
acc_{box} = \frac{1}{N_{se}}\sum_{i=1}^{N_{se}}IoU_i,
\end{equation}
where $N_{se}$ represents the number of samples in each epoch and $IoU_i$ refers to the $IoU$ of the $i$-th sample. Finally, the accuracy is a weighted sum of both the former presented components such as
\begin{equation}
    acc = w_{acc} acc_{box} + (1-w_{acc}) acc_{obj}, 
\end{equation}
where $w_{acc} \in [0,1]$ is an hyperparameter that weights the accuracy components.

\subsubsection{Real Experiments}
To assess the model's resilience, it underwent testing over distinct landing trials, each representing a specific challenge inherent to the detection scenario: modality failure and weather restrictions. In addition, the detector's inference time is measured in a proper edge computer platform to validate its viability for embedded deployment.

\begin{figure}[htbp]
\centering
\includegraphics[width=0.90\linewidth]{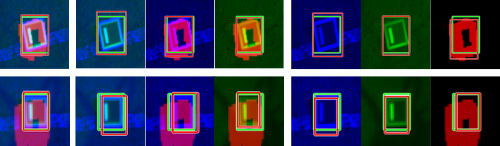}
\caption{The final representations after modality failure. In the first column, there are two examples of representations where all modalities are activated. In the second, third, and fourth columns there are examples of a disabled LiDAR, thermal and visual sensors, respectively. In the fifth, sixth and seventh column there are examples of two disabled sensors such as LiDAR and thermal, LiDAR and visual, and thermal and visual sensors, respectively.}
\label{fig:modality_failure_results}
\end{figure}

\paragraph{Modality failure test}To validate the capacity of our model to face individual sensor failure, several tests were conducted where different combinations of the three sensor modalities were disabled. The results, presented in Table \ref{tab:ablation_results}, demonstrate ViTAL's ability to achieve optimal precision and near-optimal recall scores, except when relying solely on the visual modality. The diminished performance in the visual modality can be attributed to the increased informational load, including heightened background noise that distracts the model, leading to a higher production of false negatives and consequently lower recall. The results indicate an improvement in performance with the activation of multiple modalities compared to a singular one. Notably, the mean average precision ($AP_{50}$) is higher when the visual modality is absent. This aspect is specially apparent when two modalities are active, reflecting lower AP values attributed to decreased precision relative to the unbalanced altitude distribution discussed earlier. Examples of modality failure samples disabling one modality and two modalities are shown in the Figure \ref{fig:modality_failure_results}.

\begin{table}[htbp]
\setlength{\tabcolsep}{1pt} 
    \centering
    \small
    \caption{The modality failure results. Each entry is the measured performance on the activated sensors denoted in the first column.}
    \label{tab:ablation_results}
    \begin{tabular}{c|c|c|c|c|c}
         \textbf{Active} & TPR & Recall & F1-Score & $AP_{50}$ & $AP_{50:95}$\\
         \hline
         LiDAR & 1.0 & 1.0 & 0.99 & 0.72 & 0.22\\
         Thermal & 1.0 & 0.98 & 0.99 & 0.66 & 0.16\\
         Visual & 1.0 & 0.71 & 0.83 & 0.68 & 0.16\\
         LiDAR+Thermal & 1.0 & 1.0 & 1.0 & 0.93 & 0.30\\
         Thermal+Visual & 1.0 & 1.0 & 1.0 & 0.77 & 0.40\\
         LiDAR+Visual & 1.0 & 1.0 & 1.0 & 0.82 & 0.39\\
         \bottomrule
         All & 1.0 & 1.0 & 1.0 & 0.90 & 0.45\\
    \end{tabular}
\end{table}

\begin{figure}[htbp]
\centering
\includegraphics[width=\linewidth]{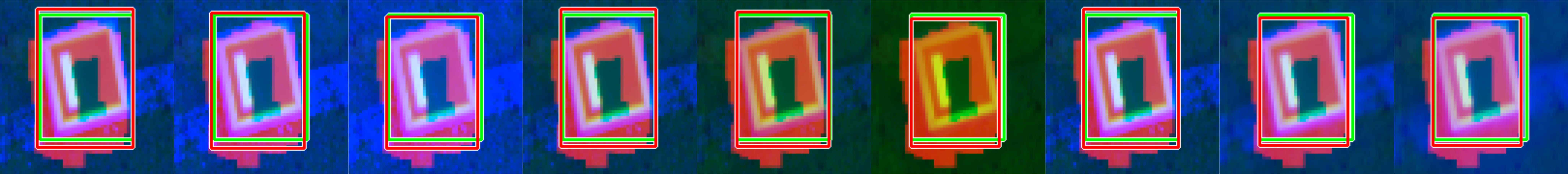}
\caption{The detection results upon weather restricted samples. The first three columns represent, from left to right, lighting variations of +10\% , +50\%, and +90\% of the image intensity. This effect is manifested through the increased intensity of the visual (blue) channel. The fourth, fifth, and sixth columns represent, from left to right, lighting variations of -10\% , -50\%, and -90\% of the image intensity. This effect is manifested through the dimmering of the visual (blue) channel. The last three columns represent, from left to right, stochastic fog effect of $[10\%,50\%]$, $[50\%,90\%]$, and $[90\%,100\%]$.}
\label{fig:weather_results}
\end{figure}

\paragraph{Weather restrictions test} To validate resilience against adverse weather, the model underwent the following conditions: lighting variations of $\pm10\%$, $\pm50\%$, and $\pm90\%$, and stochastic fog conditions of $[10\%,50\%]$, $[50\%,90\%]$, and $[90\%,100\%]$. These adversities were chosen because they are prominent in extreme weather. Brightness variations predominantly affect the visual component, while fog conditions influence all modalities. Increased brightness variations lower the model's performance by intensifying the visual component. Conversely, darkening the visual component reduces noise-induced confusion, modestly enhancing the model's performance. This slight improvement underscores the visual component's slight detriment to overall AP. Furthermore, fog restrictions have no discernible effect on performance. Overall, weather restrictions seem to not be critical to ViTAL's performance. These results are detailed in Table \ref{tab:weather_results}. Examples of weather restrictions samples for increased brightness, decreased brightness and stochastic fog conditions are shown in the Figure \ref{fig:weather_results}.

\begin{table}[htbp]
\setlength{\tabcolsep}{1pt} 
    \centering
    \small
    \caption{The weather test results. Each entry is the restriction applied to the input image received.}
    \label{tab:weather_results}
    \begin{tabular}{c|c|c|c|c|c}
         \textbf{Restriction} & TPR & Recall & F1-Score & $AP_{50}$ & $AP_{50:95}$\\ 
         \hline
         Bright $10\%$ & 1.0 & 1.0 & 1.0 & 0.90 & 0.45\\
         Bright $50\%$ & 1.0 & 1.0 & 1.0 & 0.90 & 0.42\\
         Bright $90\%$ & 1.0 & 1.0 & 1.0 & 0.90 & 0.38\\
         Dark $10\%$ & 1.0 & 1.0 & 1.0 & 0.92 & 0.46\\
         Dark $50\%$ & 1.0 & 1.0 & 1.0 & 0.82 & 0.40\\
         Dark $90\%$ & 1.0 & 1.0 & 1.0 & 0.83 & 0.30\\
         Fog $[10\%,50\%]$ & 1.0 & 1.0 & 1.0 & 0.88 & 0.44\\
         Fog $[50\%,90\%]$ & 1.0 & 1.0 & 1.0 & 0.85 & 0.43\\
         Fog $[90\%,100\%]$ & 1.0 & 1.0 & 1.0 & 0.87 & 0.40\\
         \bottomrule
         No restrictions & 1.0 & 1.0 & 1.0 & 0.90 & 0.45\\
    \end{tabular}
\end{table}

\paragraph{Inference time test} The embedded platform is a 64GB Jetson AGX Orin with configurable power deliveries of 15W, 30W, and 60W. The Table \ref{tab:detector_inference_time} displays the detector's inference times on different power loads validating its applicability for real-time operation with the worst case being approximately 50Hz. In addition, Table \ref{tab:comparison_sota_models} presents comparable performances on similar state-of-the-art models such as the DETR \cite{detr} and the ViT-B/16\cite{dosovitskiy2020image}. ViTAL surpasses both models in its inference time, which makes it substantially preferable for edge AI. Despite a less expressive performance when compared with DETR on the $AP$ metrics, DETR is far slower which makes it impracticable for a real-time application with such a high inference time of approximately 3 seconds. Nevertheless, ViTAL surpasses its most comparable architecture, the ViT-B/16, across all metrics, indicating its advanced detection capabilities. 
\begin{table}[htbp]
    \centering
    \small
    \caption{The detector's inference time for different power modes on input shapes of (3,160,160), where 3 is the color dimension, and 160 is the width and height dimension.}
    \label{tab:detector_inference_time}
    \begin{tabular}{r|c|c|c}
         \textbf{Power} (W) & 15 & 30 & 50 \\
         \hline
         \textbf{Time} (ms) & 20.07 & 8.68 & 4.97\\
    \end{tabular}
\end{table}

\begin{table}[htbp]
    \setlength{\tabcolsep}{1pt}
    \centering
    \small
    \caption{Comparison with similar state-of-the-art models. All models were trained with a batch size of 16 and a base learning rate of $3\times10^{-4}$.}
    \label{tab:comparison_sota_models}
    \begin{tabular}{c|c|c|c|c|c}
         Model & F1-Score & $AP_{50}$ & $AP_{50:95}$ & Time (ms) & $\#$ of parameters\\
         \hline
         ViT-B/16 & 0.99 & 0.07 & 0.015 &  5.20 & 86.6M\\
         DETR & 0.97 & 0.98 & 0.594 & 2707 & 39.8M \\
         \textbf{ViTAL} & 1.0 & 0.90 & 0.45 & \textbf{4.97} & 10.8M \\
    \end{tabular}
\end{table}

\subsection{Reinforcement Learning Decision-Making Lander}\label{subsec:rl_lander_results}

\subsubsection{Training}
\begin{figure}[htbp]
\centering
\includegraphics[width=0.75\linewidth]{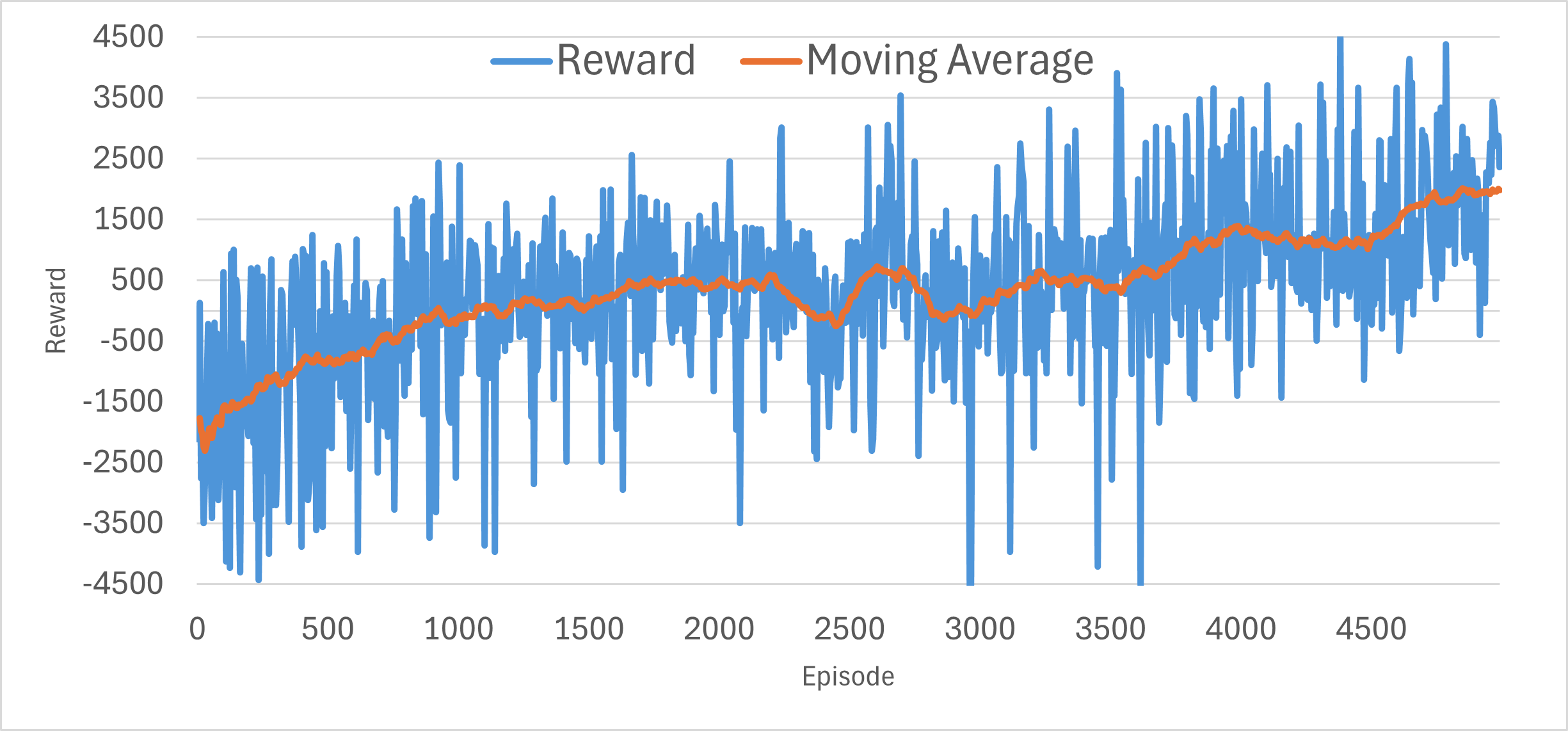}
\caption{Episode reward and moving average over episodes.}
\label{fig:reward_training}
\end{figure}

The performance of the proposed RL algorithm was evaluated through simulation tests conducted in a Gazebo environment. The experiments aimed to assess the effectiveness of the RL model in precise landing. The convergence of the RL algorithm was analyzed based on the reward function across training episodes. Figure \ref{fig:reward_training} depicts the reward and moving average obtained during training for 1.0 m grid resolution.

The plot illustrates the convergence of the RL algorithm as the mean reward increases with training episodes. The increasing trend signifies the efficacy of the learning process, showing the agent's gradual improvement in performance.

\subsubsection{Real experiments}


To assess the RL model's performance, experimental tests \footnote{The landing trials were recorded and are available at \url{https://youtu.be/ORY2JjwTIWA}} were conducted in an outdoor environment with wind conditions reaching 4m/s. The tests involved executing landing trajectories from various initial positions in a three-dimensional space. The main objective was to validate the RL algorithm's performance in real-world scenarios and confirm its ability to adapt to various initial conditions, even in the presence of wind disturbances.

\begin{figure}[htp]

\centering
\includegraphics[trim={0 0 11cm 0},clip,width=.33\linewidth]{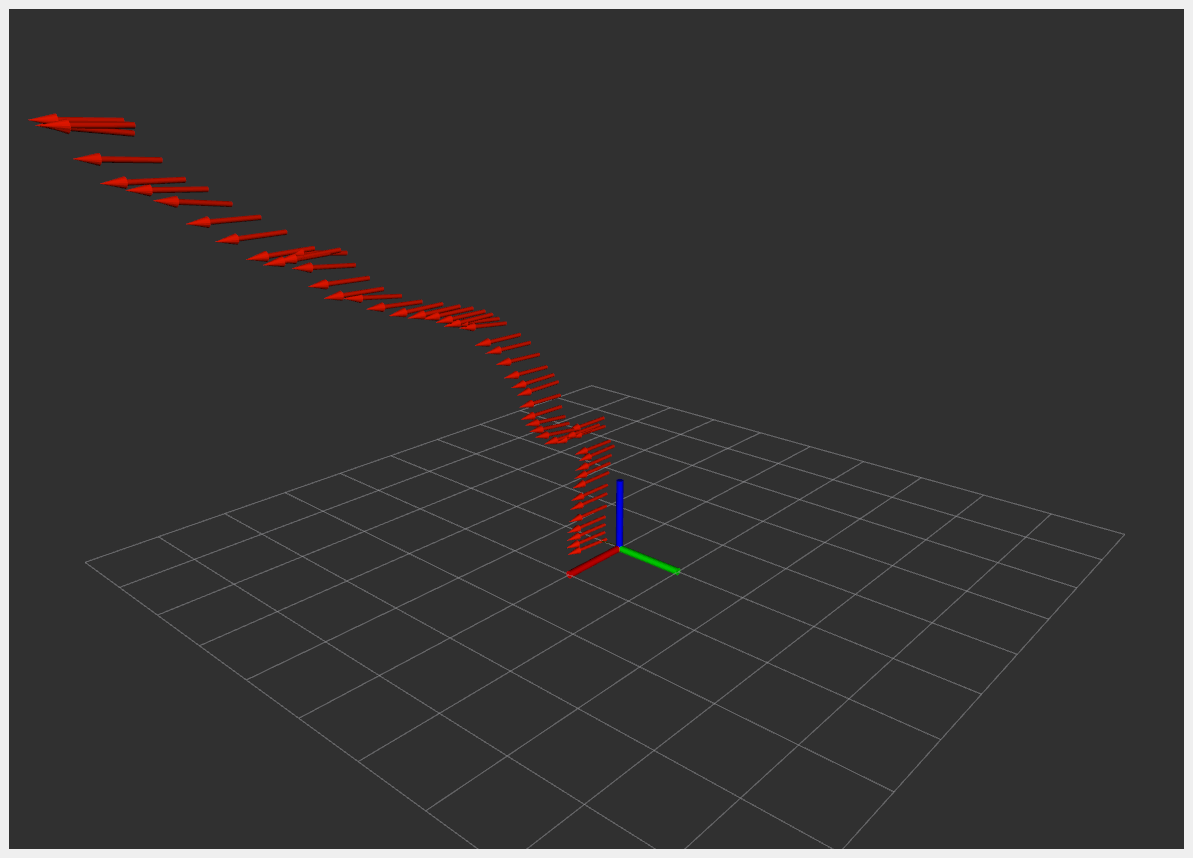}\hfill
\includegraphics[trim={1cm 0 10cm 0},clip,width=.33\linewidth]{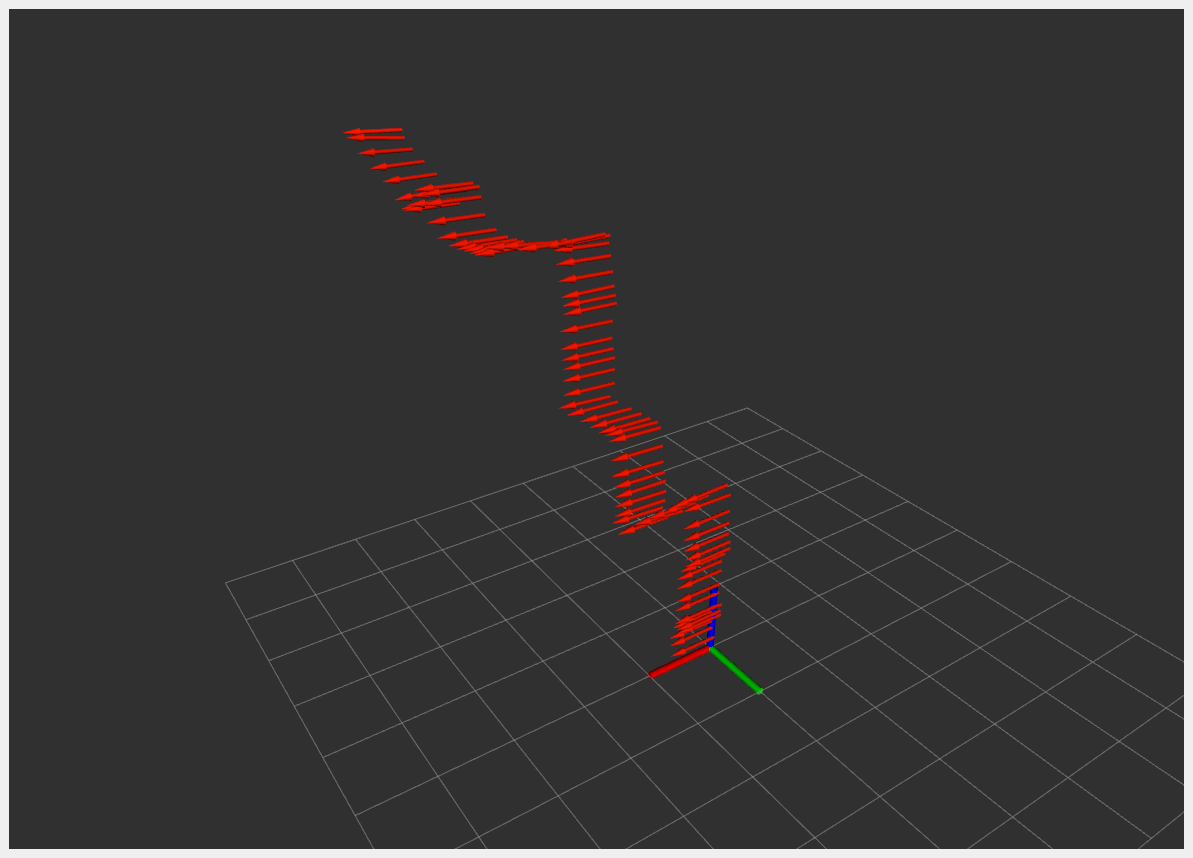}\hfill
\includegraphics[trim={6cm 0 5cm 0},clip,width=.33\linewidth]{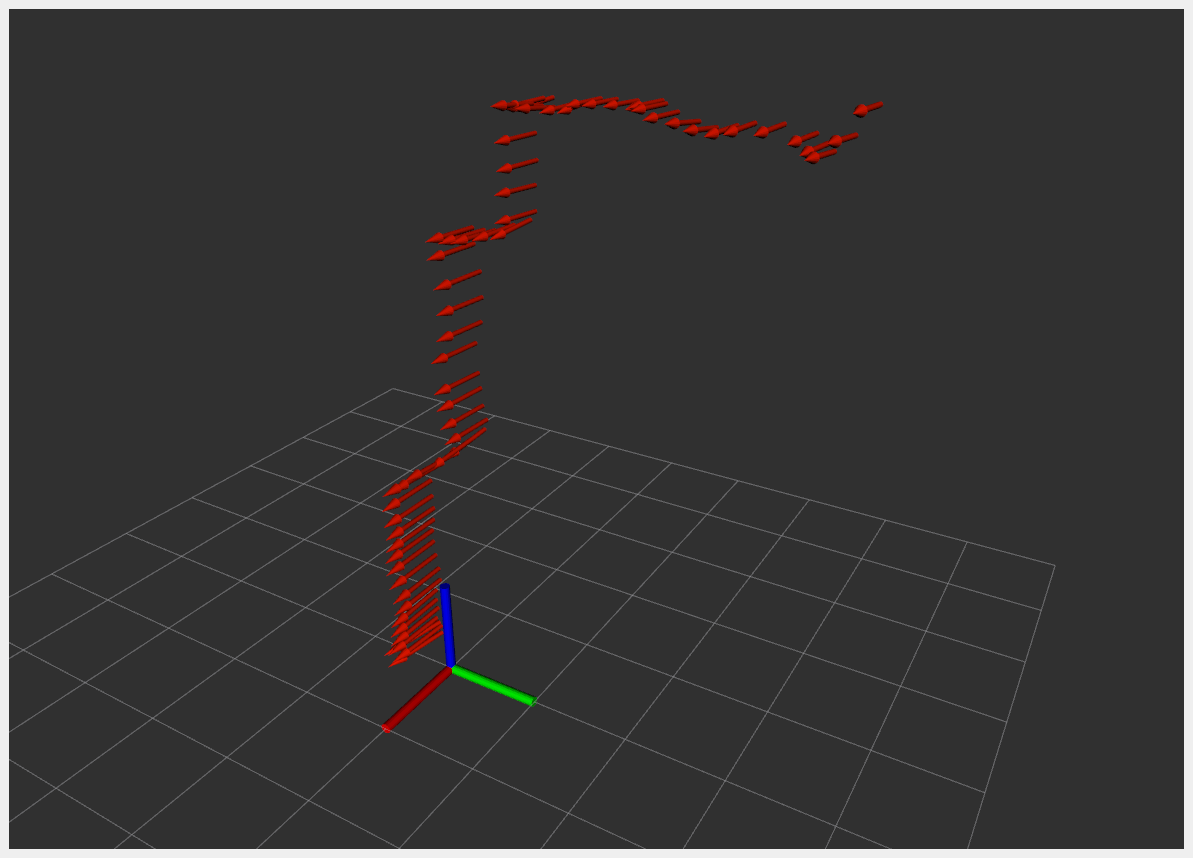}

\caption{UAV pose in landing trajectory in 3 landings.}
\label{fig:rviz_land}

\end{figure}

More precisely, the landing trajectories on three different trials with different starting positions are demonstrated in Figure \ref{fig:rviz_land}, where the red vectors refer to the pose of the UAV. Accordingly, Table \ref{tab:precise_error_landing} presents the corresponding landing deviations to the center of the landing pad. It can be concluded that there was a successful precise landing accuracy with an average value of 25 centimeters relative to the center of the landing pad.

\begin{table}[htbp]
    \centering
    \small
    \caption{Landing deviation relative to the center of the landing pad for three different trials, with a sensor's positioning accuracy of 0.025m.}
    \label{tab:precise_error_landing}
    \begin{tabular}{r|c|c|c}
         \textbf{\# Trial} & 1 & 2 & 3 \\
         \hline
         \textbf{Deviation} (m) & 0.36 & 0.03 & 0.38\\
    \end{tabular}
\end{table}

The results demonstrate the advantage of a learning-based approach in generalizing to diverse situations, compared to complex mathematical methods that may prove too limiting in unpredictable environments. This validation process underscored the algorithm's resilience to potential disturbances, highlighting its effectiveness in addressing complex real-world challenges.

\section{Conclusions}\label{sec:conclusions}
This research presents an innovative autonomous landing approach for UAVs that relies on a multimodal-based landing target detector an autonomous landing navigational approach landing based on reinforcement learning. It is an integrated reasoning framework benefiting from learning-based perception and decision-making modules.

On one hand, it is delivered resilient perception capabilities to detect the landing platform due to its inherent multimodal nature. On the other hand, by availing from an increasingly reliable input, the decision-making lander, can operate in a more continuous and effective manner. Besides, the detected position has centimeter-level precision which surpasses by a couple orders of magnitude the traditional positioning of GPS. Nonetheless, this system's resiliency is in part significantly linked with such a multimodal property, since the full capability is exploited if the UAV is equipped with heterogeneous sensors capable of detecting multimodal markers.

Both modules were validated in real settings, showcasing its robotic applicability. The system was also inferred on an edge device demonstrating real-time operability with average inference times up to 200 FPS. Moreover, and more precisely, the detector ViTAL achieved full operability on different weather and modality failure restrictions, which can be observed by its consistent optimal TPR value. In addition, the RL module has demonstrated the benefits of choosing a learning-based approach to address complex real-world challenges. Its convergence was evident through the increasing moving average of the reward during the training phase, coupled with successful simulation tests showcasing autonomous landing maneuvers. The UAV's behavior aligns with the formulated reward functions: initially directed towards the landing zone, followed by a transition to a nearly vertical descent leading to landing on the designated pad. The real experiments culminate on a successful precise landing with an average accuracy of 0.25 m relative to the center of the landing pad.





\vspace{12pt}
\end{document}